\documentclass[conference]{IEEEtran}
\IEEEoverridecommandlockouts
\usepackage{cite}
\usepackage{amsmath,amssymb,amsfonts}
\usepackage{algorithmic}
\usepackage{graphicx}
\usepackage
{textcomp}
\usepackage{xcolor}
\def\BibTeX{{\rm B\kern-.05em{\sc i\kern-.025em b}\kern-.08em
    T\kern-.1667em\lower.7ex\hbox{E}\kern-.125emX}}

\usepackage{url}
\usepackage{multirow}
\usepackage{siunitx}
\usepackage{bm}

\usepackage[switch]{lineno}

\begin{document}

\title{Depth Estimation fusing Image and\\ Radar Measurements with Uncertain Directions
}

\author{\IEEEauthorblockN{Anonymous Authors}}

\author{\IEEEauthorblockN{Masaya Kotani}
\IEEEauthorblockA{\textit{Toyota Technological Institute}, Japan \\
masaya.kotani.ttij@gmail.com}
\and
\IEEEauthorblockN{Takeru Oba}
\IEEEauthorblockA{\textit{Toyota Technological Institute}, Japan \\
sd21502@toyota-ti.ac.jp}
\and
\IEEEauthorblockN{Norimichi Ukita}
\IEEEauthorblockA{\textit{Toyota Technological Institute}, Japan \\
ukita@toyota-ti.ac.jp}
}

\maketitle

\begin{abstract}
This paper proposes a depth estimation method using radar-image fusion by addressing the uncertain vertical directions of sparse radar measurements.
In prior radar-image fusion work, image features are merged with the uncertain sparse depths measured by radar through convolutional layers.
This approach is disturbed by the features computed with the uncertain radar depths.
Furthermore, since the features are computed with a fully convolutional network, the uncertainty of each depth corresponding to a pixel is spread out over its surrounding pixels.
Our method avoids this problem by computing features only with an image and conditioning the features pixelwise with the radar depth.
Furthermore, the set of possibly correct radar directions is identified with reliable LiDAR measurements, which are available only in the training stage.
Our method improves training data by learning only these possibly correct radar directions, while the previous method trains raw radar measurements, including erroneous measurements.
Experimental results demonstrate that our method can improve the quantitative and qualitative results compared with its base method using radar-image fusion.
\end{abstract}



\section{Introduction}
\label{section:introduction}

Perception of vehicle surroundings is essential for self-driving and ADAS (Advanced Driver-Assistance Systems).
For such perception (e.g., object detection~\cite{DBLP:journals/tits/FengHRHGTWD21,DBLP:conf/wacv/NabatiQ21} and depth estimation~\cite{DBLP:journals/corr/abs-2205-05335}), a variety of sensors are useful.
RGB cameras and LiDAR are typical for such sensors because they can capture a wide range.
However, their measurements are disturbed by bad weather such as fog and rain because they capture short wavelengths (e.g., \SIrange{400}{700}{\nm} for RGB and \SIrange{0.4}{1.5}{\um} for LiDAR).
In addition, LiDAR is not common for products yet due to its high price.
On the other hand, a millimeter-wave radar can work without being disturbed by bad weather because of its long wavelength measurement (i.e., \SIrange{1.0}{10.0}{\mm}).
While highly functional millimeter-wave radars such as scanning radar~\cite{scanradar} and high-resolution radar~\cite{DBLP:conf/cvpr/RebutOMP22} are also developed, these novel radars are still expensive, voluminous, and low temporal resolution.
For the reasons mentioned above, a combination of a camera and a general millimeter-wave radar is one of the good choices for several on-vehicle vision systems such as object detection~\cite{DBLP:conf/sdf/NobisGWBL19} and depth estimation~\cite{DBLP:conf/cvpr/Long00CCN21}.

\begin{figure}[t]
  \begin{center}
    \includegraphics[width=\columnwidth]{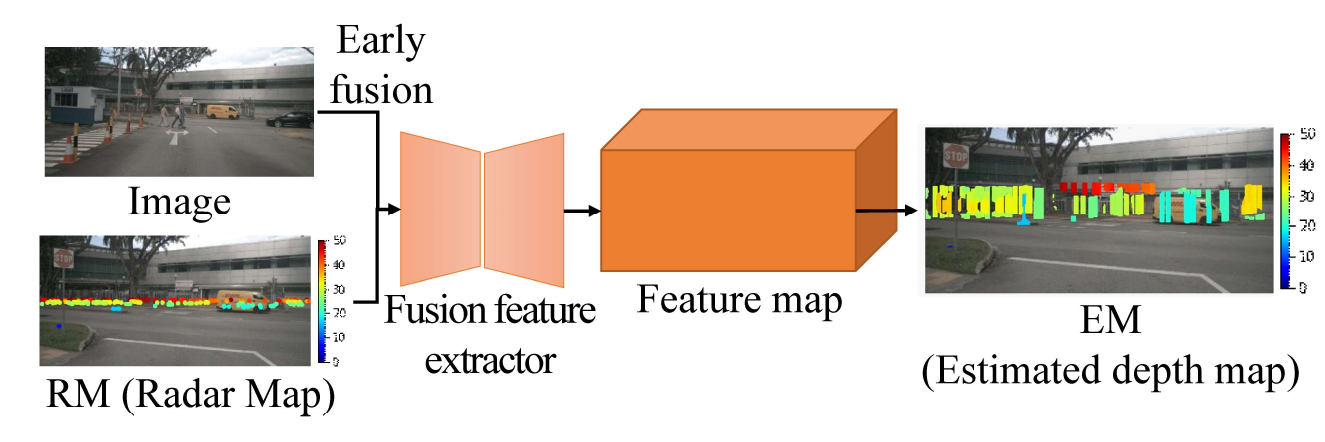}\\
    \vspace*{-1mm}
    (Top) Previous method~\cite{DBLP:conf/cvpr/Long00CCN21}\\
    \vspace*{7mm}
    \includegraphics[width=\columnwidth]{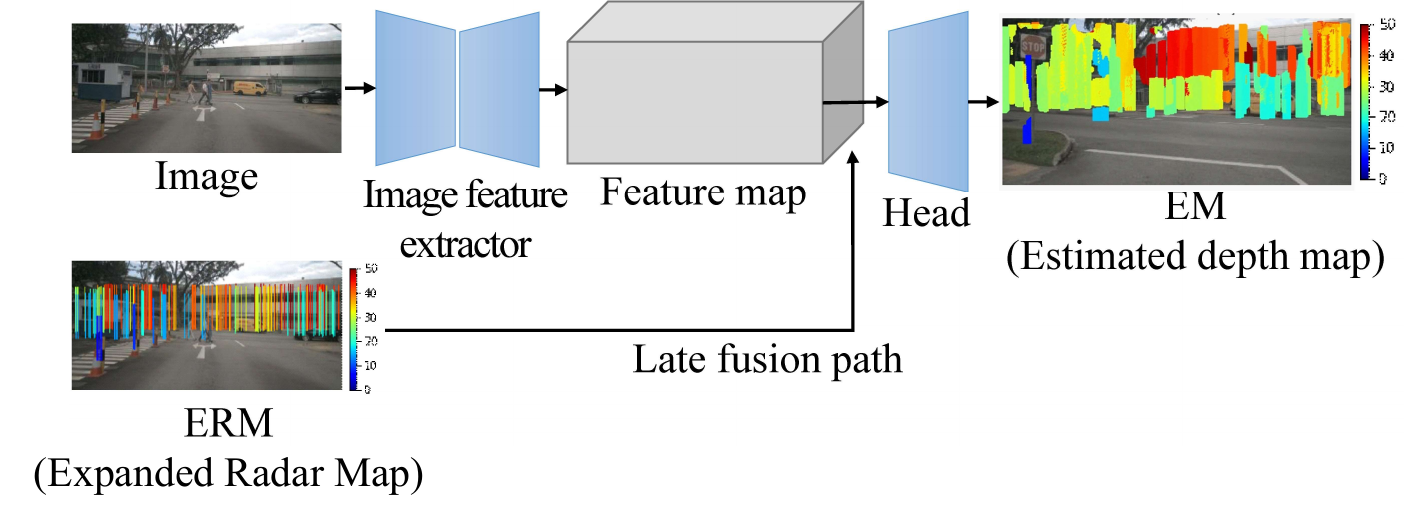}
    \vspace*{-1mm}
    (Bottom) Our proposed method
  \end{center}
  \caption{Difference between our method with late radar fusion (Bottom) and its closest work~\cite{DBLP:conf/cvpr/Long00CCN21} with early radar fusion (Top).
   In the network shown at the top, erroneous radar measurements are merged with an input image in the feature extraction process. To avoid such image feature contamination, in our proposed network shown at the bottom, image feature extraction is separated from the image-radar fusion process.
   While auxiliary image-based cues are also fed into the feature extractor both in the previous and our methods, these cues are omitted for brevity.
   All depth maps, such as RM, ERM, and EM, are overlaid on images in all figures for visualization in this paper.
  }
  \label{fig:conventional_vs_proposed}
\end{figure}

However, the most significant disadvantage of a general millimeter-wave radar, called radar for brevity in what follows, is that its measurement directions are reliable along the horizontal direction but uncertain along the vertical direction.
This uncertainty makes it difficult to align the measured radar points on an image spatially.
In Long et al.~\cite{DBLP:conf/cvpr/Long00CCN21}, for example, the radar points are projected onto the image plane, assuming that all the radar measurement directions are parallel to the horizontal plane, as shown in the upper input image (i.e., ``RM'' in the figure) of the upper network in Fig.~\ref{fig:conventional_vs_proposed}.
Such horizontal projection is not always correct due to the uncertainty of the vertical directions of the radar measurements.
Since these incorrect radar projections are fed into a feature extractor (e.g., convolutional network) with an image based on an early fusion manner, as shown in Fig.~\ref{fig:conventional_vs_proposed} (Top), the incorrect radar information is spread out to features extracted by the feature extractor.

\begin{figure*}[t]
  \begin{center}
     \includegraphics[width=0.80\linewidth]{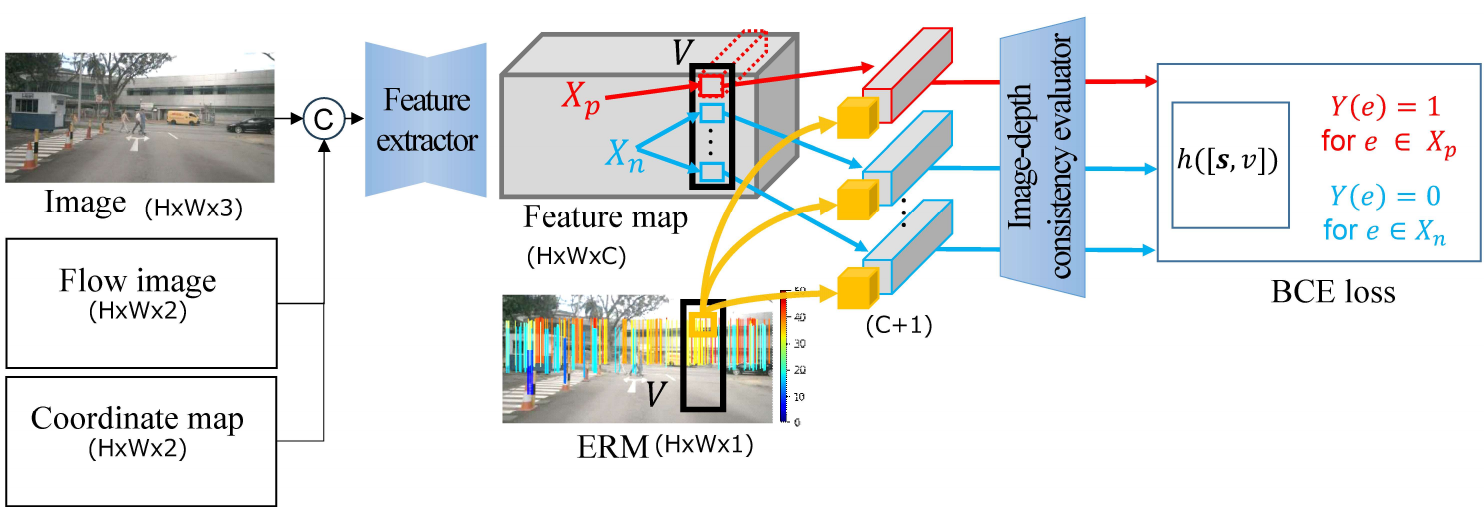}\\
     \vspace*{-2mm}
     (Top) Training procedure\\
     \vspace*{2mm}
     \includegraphics[width=0.80\linewidth]{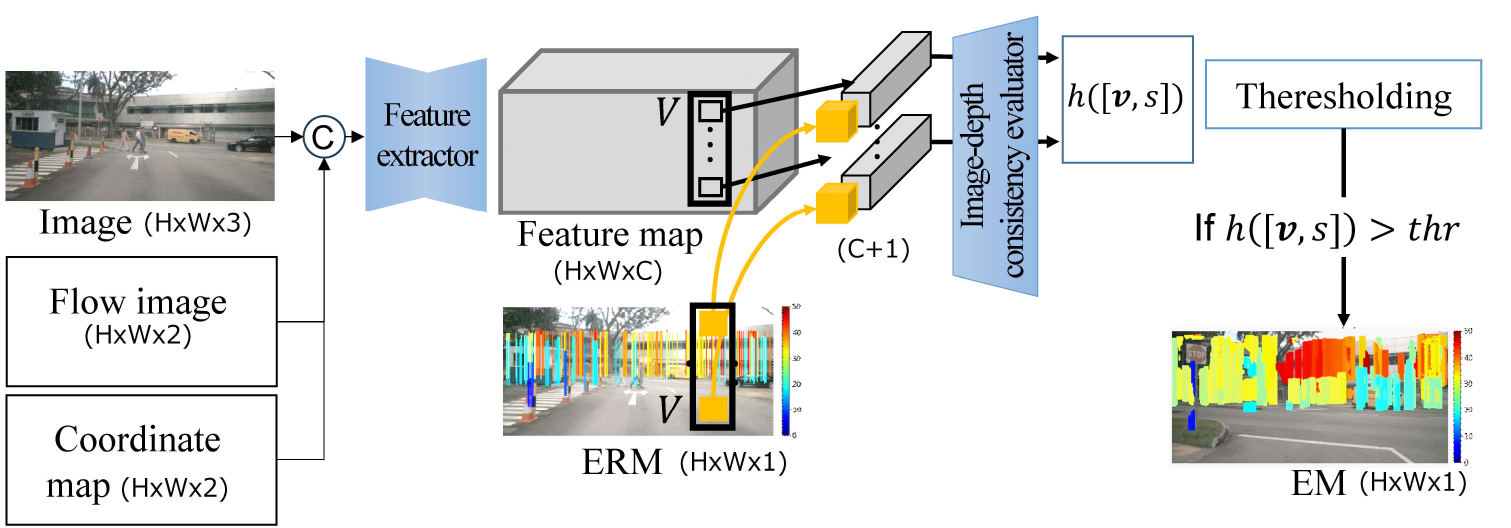}\\
     \vspace*{-2mm}
     (Bottom) Inference procedure
  \end{center}
     \vspace*{-2mm}
  \caption{Full pipeline of our method. Top: Training. Bottom: Inference. The channel size of each data is indicated within the parentheses.}
  \label{fig:method}
  \vspace*{-2mm}
\end{figure*}

This paper aims to suppress the problem of uncertain radar directions with the contributions below:
\begin{itemize}
    \item Image features are extracted without being disturbed by uncertain radar measurements by feeding only an image into a feature extraction network.
    The radar points are concatenated pixelwise with the extracted features in a late fusion manner for supporting depth estimation from the image features.
    \item In training, the possibly correct radar directions are identified by supervision with LiDAR measurements.
    The features of a pixel corresponding to each possibly correct radar direction are paired with the radar-depth value as a correct image-radar pair. This pairing is done via the late fusion path, as shown in the bottom network in Fig.~\ref{fig:conventional_vs_proposed}.
    \item By training a set of correct image-radar pairs, our method achieves pixelwise depth estimation without being interfered with by erroneous radar measurements.
\end{itemize}


\section{Related Work}

\paragraph{Radar-RGB Fusion}

In~\cite{DBLP:conf/icra/ChadwickM019}, image and radar features are merged in a late fusion manner to improve the object detection performance in contrast to using only image features.
However, the direct use of uncertain radar points degrades the performance, as validated in~\cite{DBLP:conf/iros/LinDG20}.

To cope with the uncertainty of radar directions, each radar measurement is expressed not as a point but as a line segment in an image plane for image-radar early fusion in~\cite{DBLP:conf/sdf/NobisGWBL19}.
In addition to this scheme, in this method~\cite{DBLP:conf/sdf/NobisGWBL19}, the priority of image feature training is decreased after appropriate training iterations to focus on radar feature training because the data amount of measured radar points is significantly less than the image information (i.e., image pixels).
As well as the vertical uncertainty, the small horizontal uncertainty is addressed in~\cite{DBLP:conf/wacv/StackerHRS22}.
While these methods~\cite{DBLP:conf/sdf/NobisGWBL19,DBLP:conf/wacv/StackerHRS22} extend the radar points to other representations to avoid the direct use of uncertain radar points, the uncertain information is spread out by convolutions due to early fusion in these methods.

Different from the aforementioned radar augmentation approaches~\cite{DBLP:conf/sdf/NobisGWBL19,DBLP:conf/wacv/StackerHRS22}, several previous methods focus on how to reduce the negative impact of the uncertain radar points.
In~\cite{DBLP:conf/iros/LinDG20}, image-radar fusion features obtained by early fusion are further conditioned by uncertain radar points to learn the negative impacts of the uncertain radar points on the extracted features.
The effectiveness of network pruning for image-radar fusion, which is expected to help reduce the negative impact of the uncertain radar points, is validated in~\cite{DBLP:conf/wacv/StackerHRS22}.
In~\cite{DBLP:conf/cvpr/Long00CCN21}, only limited image pixels close to each radar point are related to the depth value of this radar measurement for learning the image features corresponding to (i.e., predicting) this depth value so that the uncertainty of each pixel is not spread out over the image.
While these approaches can decrease the negative impact of the uncertain radar points, these methods also apply convolutions to image-radar features so that the uncertainty of the radar directions is spread out over the features due to early fusion.


\section{Proposed Method}
\label{section:method}


\subsection{Overview}
\label{subsection:overview}

The overview of the proposed framework, whose brief overview is illustrated in Fig.~\ref{fig:conventional_vs_proposed} (Bottom), is shown in Fig.~\ref{fig:method}.
In the top and bottom sides of Fig.~\ref{fig:method}, the training and inference procedures of the network are shown, respectively.

Given an input image (i.e., ``Image'' in Fig.~\ref{fig:method}), its optical flow image and normalized coordinate map are obtained.
The optical flow image is estimated by RAFT (Recurrent All-pairs Field Transforms for optical flow)~\cite{DBLP:journals/spl/JiaWLLL21}.
In the normalized coordinate map, its $x$ and $y$ channels have the $x$ and $y$ coordinate values normalized between 0 and 1 in each pixel, respectively.

\begin{figure}[t]
  \begin{center}
  \includegraphics[width=\columnwidth]{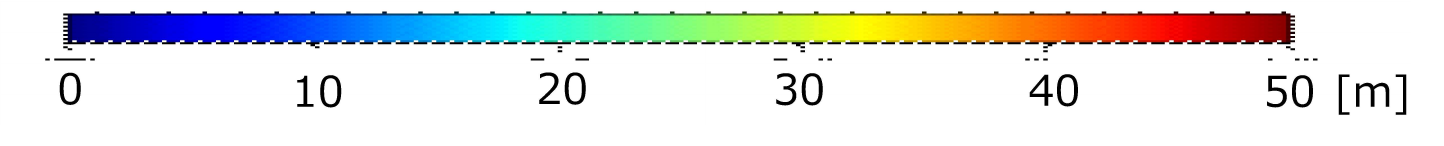}
  \begin{minipage}{0.49\columnwidth}
    \includegraphics[width=\columnwidth]{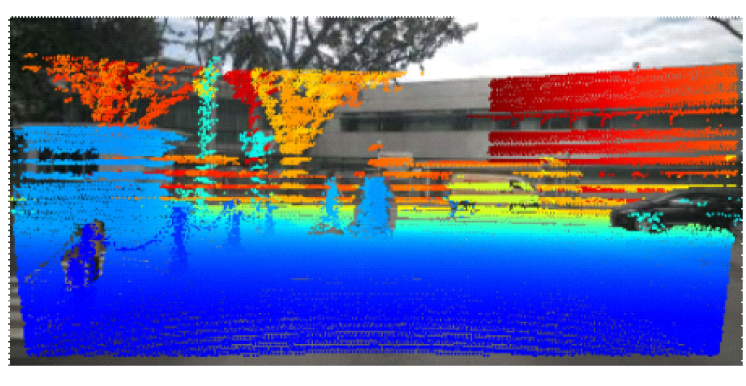}\\
    (a) LiDAR Map (LM) \\
    \vspace*{2mm}
    \includegraphics[width=\columnwidth]{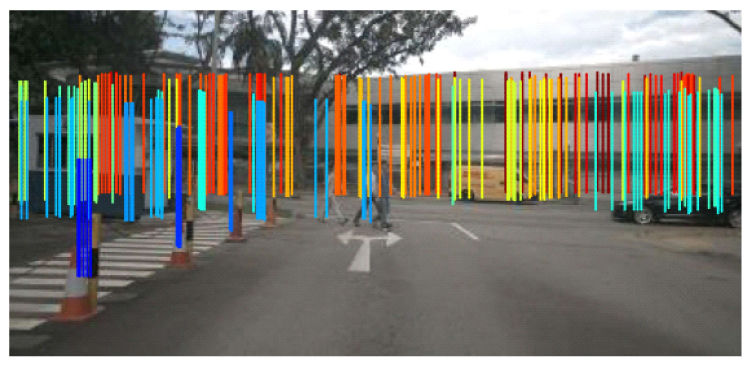}\\
    (c) Extended Radar Map (ERM)
  \end{minipage}
  \begin{minipage}{0.49\columnwidth}
    \includegraphics[width=\columnwidth]{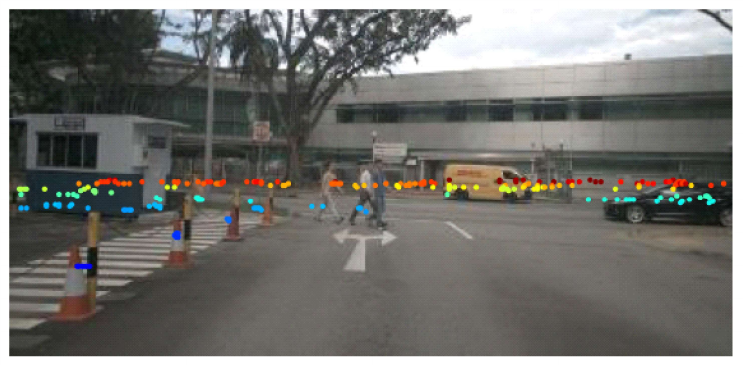}\\
    (b) Radar Map (RM)\\
    \vspace*{2mm}
    \includegraphics[width=\columnwidth]{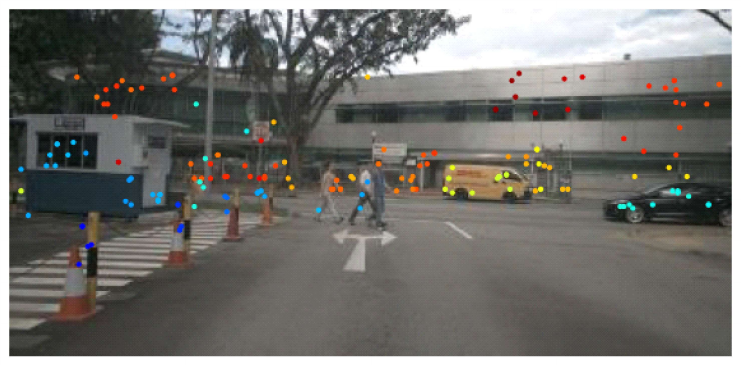}\\
    (d) Possibly correct Radar Map (PCRM)
  \end{minipage}
  \end{center}
  \caption{Depth points measured by (a) LiDAR and (b) radar, which are called LM and RM, respectively. (c) ERM: The radar points are expanded along the $y$ axis. (d) PCRM: Possibly correct radar points are selected from ERM by using LM.}
  \label{fig:depths}
\end{figure}


\subsection{Preliminary}
\label{subsection:preliminary}

In addition to the image and radar, which are available both in the training and inference procedures, LiDAR points projected onto an image plane (Fig.~\ref{fig:depths} (a)) are also employed as supervised data in the training procedure.

As mentioned before, radar points are assumed to be measured along the horizontal plane, as shown in Fig.~\ref{fig:depths} (b), in prior work.
In our method, on the other hand, the possibly correct directions of radar measurements are identified as follows.
Given the vertical measurement range of the radar device (denoted by $V$ pixels), all radar points shown in Fig.~\ref{fig:depths} (b) are expanded over $V$ pixels along the vertical axis, as shown in Fig.~\ref{fig:depths} (c).
The image coordinates of this point set are denoted by $P_{erm}=\{ \bm{p}_{erm} (e) \}_{e=1}^{RV} = \{ (x_{1}, y_{1}), \cdots, (x_{RV}, y_{RV}) \}$ where $R$ denotes the number of the radar points.
A sparse depth map in which depth values are substituted only in $\bm{p}_{erm}(e)$ where $e \in \{1,\cdots,RV\}$ is denoted by $ERM(\bm{i})$ where $\bm{i}$ denotes image $xy$ coordinates.
This sparse depth map is called an extended radar map (ERM), shown in Fig.~\ref{fig:depths} (c).

Assuming that the depth values of the radar and LiDAR are reliable (while the measurement direction of each radar point is unreliable), we select the expanded radar points, each of whose depth matches the depth of LiDAR in each pixel.
This matching is satisfied if the absolute and relative differences between the depths of the expanded radar and LiDAR points are small.
Specifically, given the depth values of radar and LiDAR denoted by $d_{r}$ and $d_{l}$, respectively, 
if $|| d_{l} - d_{r} ||$ is less than $T_{a}$ and $\frac{|| d_{l} - d_{r} ||}{d_{l}}$ is less than $T_{r}$, the matching is satisfied.
$T_{a}$ and $T_{r}$ denote the hyper-parameters for thresholding.
These matched radar points are regarded as possibly correct radar measurements composing a Possibly correct Radar Map (PCRM), which is shown in Fig.~\ref{fig:depths} (d).
The index sets of the points in which this condition is satisfied and unsatisfied are denoted by $\bm{X}_{p}$ and $\bm{X}_{n}$, respectively.


\subsection{Training}
\label{subsection:training}

The overview of the training pipeline in our method is shown in Fig.~\ref{fig:method} (Top).
The feature extraction network extracts an image feature map, $F$, from an input image with its flow image and normalized coordinate map.
The features in $xy$ coordinates $\bm{i}$ are denoted by $F(\bm{i})$.
From this feature map, image features are extracted in possibly correct radar points defined by $\bm{X}_{p}$ and possibly incorrect radar points defined by $\bm{X}_{n}$.
These extracted features are used as the supervised data for training the image-depth consistency evaluation network, $H(e)$, that judges whether or not a pair of the radar depth value and the features in $e$-th pixel of ERM is probable.
The pairs are probable and not probable in $\bm{X}_{p}$ and $\bm{X}_{n}$, respectively.
The aforementioned process is formulated so that $H(e) = h([\bm{v}, s])$ estimates a probability $p(e)$ that a depth value $s$ is observed in a pixel whose image features are $\bm{v}$, as follows:
\begin{eqnarray}
h \left( [F(\bm{p}_{erm}{(e)}), ERM(\bm{p}_{erm}(e))] \right) &=& p(e),\label{eq:PCRM}
\end{eqnarray}
where $[ \bm{v}, s ]$ denotes the concatenation of $\bm{v}$ and $s$, which are depicted by ``each gray cuboid enclosed by the red/blue line'' and ``each yellow cuboid,'' respectively, in Fig.~\ref{fig:method} (Top).
The channel sizes of $\bm{v}$ and $s$ are $c$ and $1$, respectively.
The ground truth of $p(e)$ is $Y(e)$.
$Y(e) = 1$ if $e \in \bm{X}_{p}$, otherwise (i.e., if $e \in \bm{X}_{n}$) $Y(e) = 0$.

With the positive and negative samples expressed by Eq.~(\ref{eq:PCRM}), the feature extraction network and the image-depth consistency evaluation network are trained in an end-to-end manner with the following weighted binary cross entropy loss:
\begin{equation}
\sum_{e=1}^{RV} w(e) \left( - Y(e) \log(p(e)) - (1 - Y(e)) \log(1 - p(e)) \right),
\end{equation}
where $w(e)$ is a weight parameter for coping with a class imbalance between the positive and negative samples.
\begin{equation}
w(e) = \frac{| \bm{X}_{p} |}{| \bm{X}_{p} | + | \bm{X}_{n} |},
\end{equation}
if $e$-th pixel is included in $\bm{X}_{p}$.
If $e$-th pixel is included in $\bm{X}_{n}$,
\begin{equation}
w(e) = \frac{| \bm{X}_{n} |}{| \bm{X}_{p} | + | \bm{X}_{n} |}.
\end{equation}


\subsection{Inference}
\label{subsection:inference}

The inference procedure is illustrated in Fig.~\ref{fig:method} (Bottom).
While possibly correct radar points can be identified by LiDAR points in training, the LiDAR points are not available in inference.
Therefore, image features in every pixel included in ERM are evaluated by Eq.~(\ref{eq:PCRM}) in the image-depth consistency evaluation network.
Specifically, image features $F(\bm{p}_{erm} (e))$ and depth value $ERM(\bm{p}_{erm} (e))$ are extracted from $F$ and ERM, respectively, and the concatenation of $F(\bm{p}_{erm} (e))$ and $ERM(\bm{p}_{erm} (e))$ are fed into the the image-depth consistency evaluation network.

If the output from the image-depth consistency evaluation network, $h \left( [F(\bm{p}_{erm}{(e)}), ERM(\bm{p}_{erm}(e))] \right)$ depicted by ``$h([\bm{v},s])$'' in Fig.~\ref{fig:method}, is greater than a threshold, $ERM(\bm{p}_{erm} (e))$ is substituted to the depth value in $e$-th pixel of the estimated depth map (i.e., ``EM'' in Fig.~\ref{fig:method} (Bottom)).
Otherwise, the depth value in $e$-th pixel of EM is regarded as unavailable.

After the aforementioned depth substitution process for all pixels in ERM, the estimated depth map, EM, is acquired.

Since the depth value is not available in many pixels in EM, all depth values are estimated by a depth completion method~\cite{DBLP:conf/icra/MaK18,DBLP:conf/wacv/LiYLCZZ20}.


\subsection{Network Architectures}

\begin{figure}[t]
  \begin{center}
     \includegraphics[width=\linewidth]{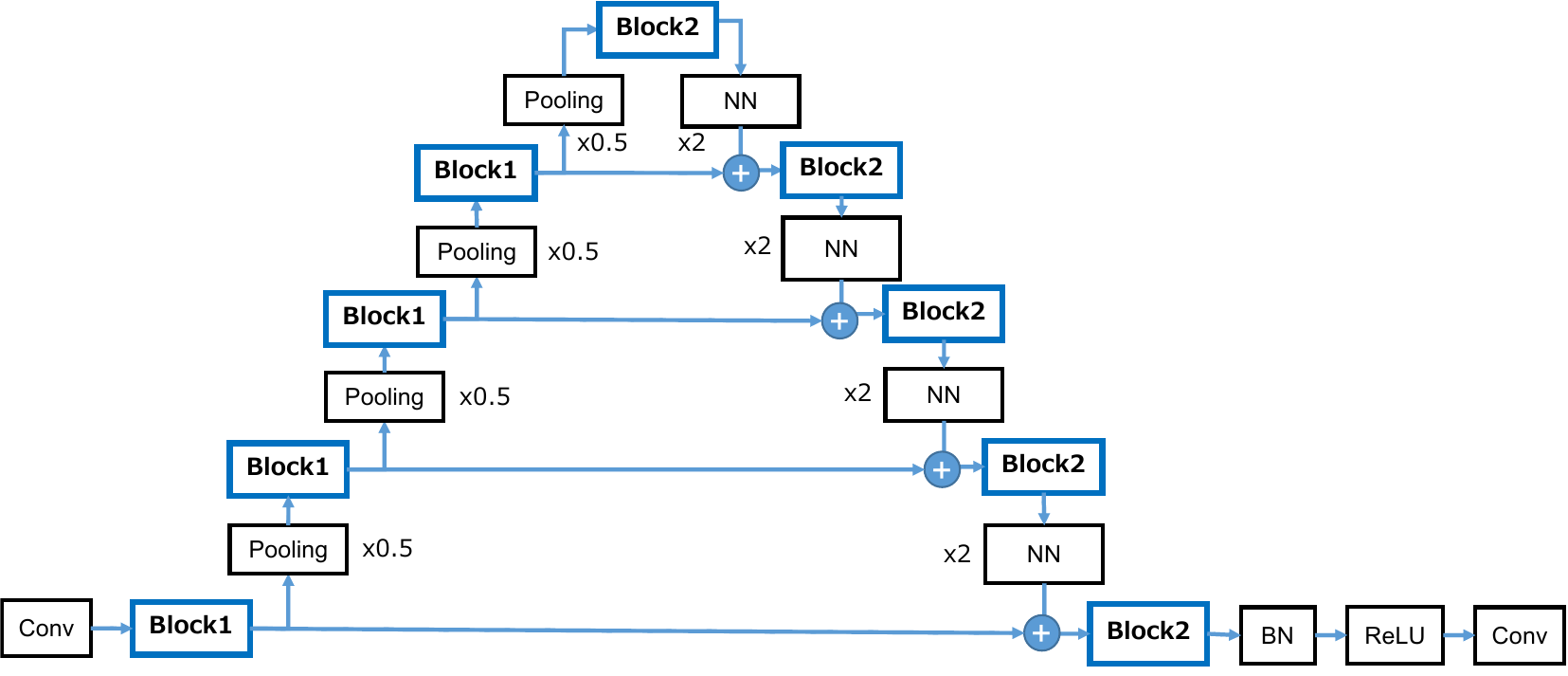}\\
     Feature Extraction Network\\
     \vspace*{5mm}
     \includegraphics[width=\linewidth]{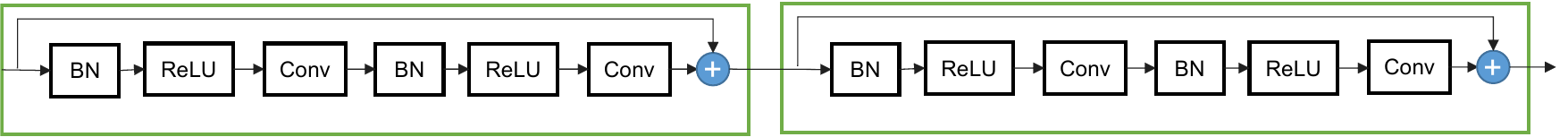}\\
     Block1\\
     \vspace*{5mm}
     \includegraphics[width=\linewidth]{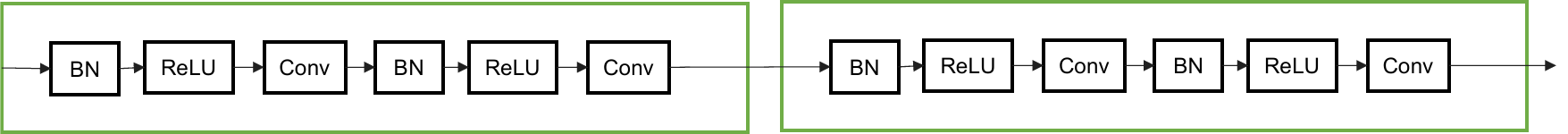}\\
     Block2
  \end{center}
  \caption{Network architecture of the feature extraction network. Conv, Pooling, NN, BN, and ReLU denote a convolution layer, a max pooling layer for downsampling, an upsampling layer using nearest neighbor sampling, a batch normalization layer, and a rectified linear unit, respectively.}
  \label{fig:network}
\end{figure}

\begin{figure}[t]
  \begin{center}
     \includegraphics[width=0.75\linewidth]{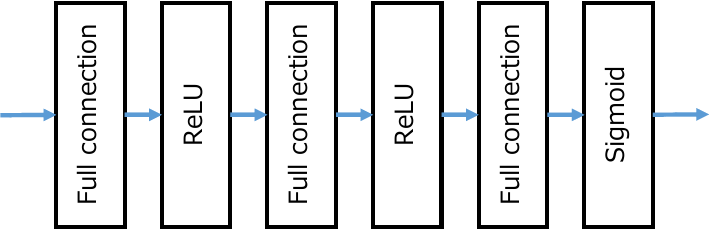}
  \end{center}
  \caption{Network architecture of the image-depth consistency evaluation network.}
  \label{fig:idce}
\end{figure}

The detailed architectures of the feature extraction network (i.e., ``Feature extractor'' in Fig.~\ref{fig:method}) and the image-depth consistency evaluation network (i.e., ``Image-depth consistency evaluator'' in Fig.~\ref{fig:method}) are shown in Fig.~\ref{fig:network} and Fig.~\ref{fig:idce}, respectively.


\section{Experimental Results}
\label{section:experiments}

\begin{table}[t]
\caption{Depth completion errors (meters). Comparison between different depth input patterns.}
\label{Tab:result1}
\centering
\begin{tabular}{lc||lll}
\hline
& Completion & \multicolumn{3}{c}{Error↓}                     \\ \cline{3-5} 
\multicolumn{1}{c}{Methods} & methods & \multicolumn{1}{c}{MAE} & \multicolumn{1}{c}{REL} & \multicolumn{1}{c}{RMSE} \\ \hline\hline
RM          & \multirow{4}{*}{Ma~\cite{DBLP:conf/icra/MaK18}} & 1.57 & 0.090 & 3.33 \\
Long~\cite{DBLP:conf/cvpr/Long00CCN21} &                          & 1.47 & 0.085 & 3.19 \\
\bf{Ours }  &                          & \bf{1.44} & \bf{0.080} & \bf{3.07} \\
PCRM &                          & 0.91 & 0.057 & 2.33 \\ \hline
RM          & \multirow{4}{*}{Li~\cite{DBLP:conf/wacv/LiYLCZZ20}} & 1.82 & 0.107 & 3.65 \\
Long~\cite{DBLP:conf/cvpr/Long00CCN21} &                          & 1.66 & 0.094 & 3.46 \\
\bf{Ours}   &                          & \bf{1.62} & \bf{0.092} & \bf{3.34} \\
PCRM &                          & 1.15 & 0.073 & 2.69 \\ \hline
\end{tabular}
\end{table}

\begin{figure*}[t]
  \begin{center}
    \includegraphics[width=0.32\linewidth]{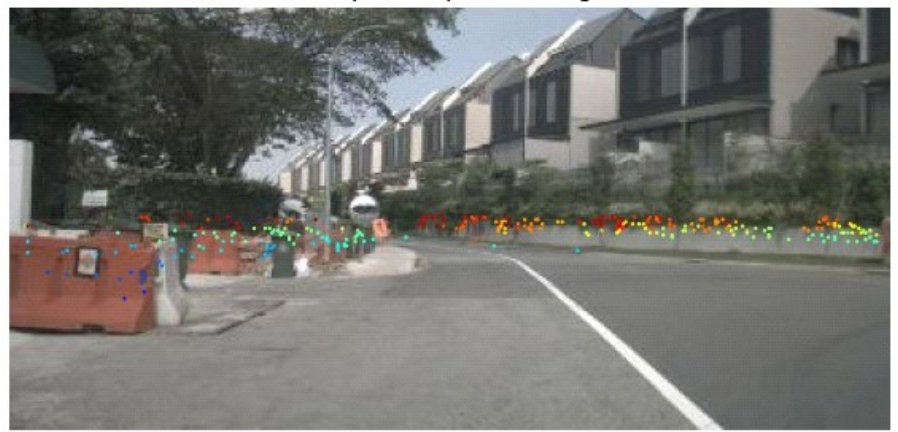}~
    \includegraphics[width=0.32\linewidth]{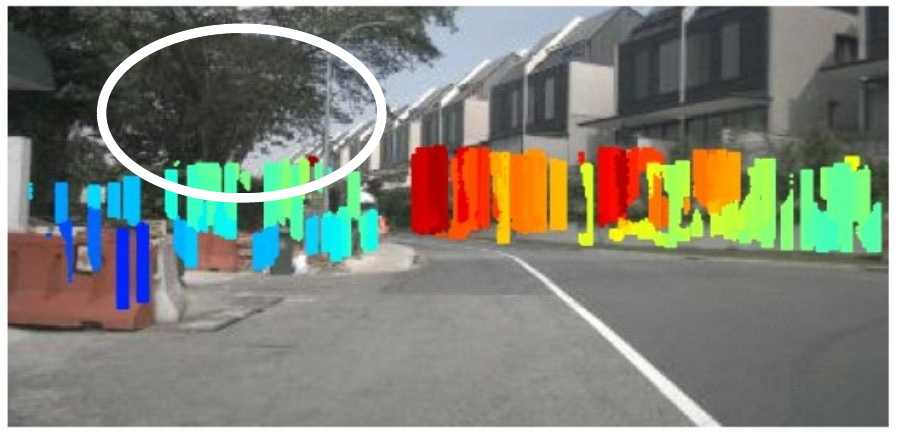}~
    \includegraphics[width=0.32\linewidth]{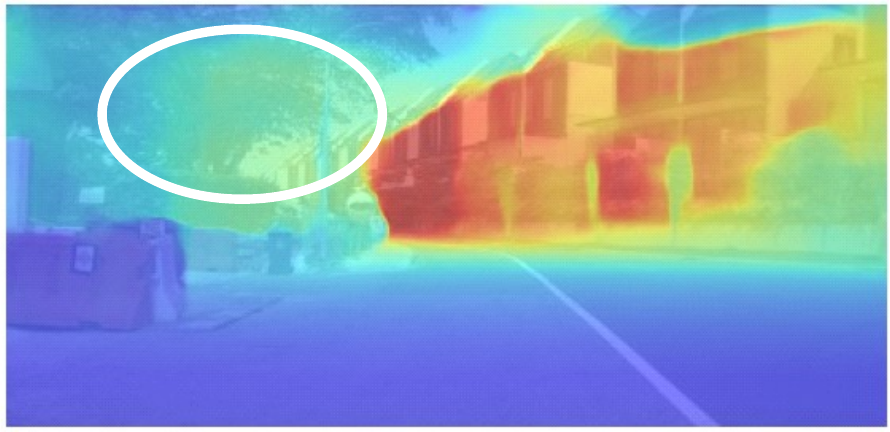}\\
    ~\hspace*{15mm} (a) RM \hspace*{45mm} (c) EM of~\cite{DBLP:conf/cvpr/Long00CCN21} \hspace*{25mm} (e) Depth map completed from (c)\\
        \vspace*{2mm}
    \includegraphics[width=0.32\linewidth]{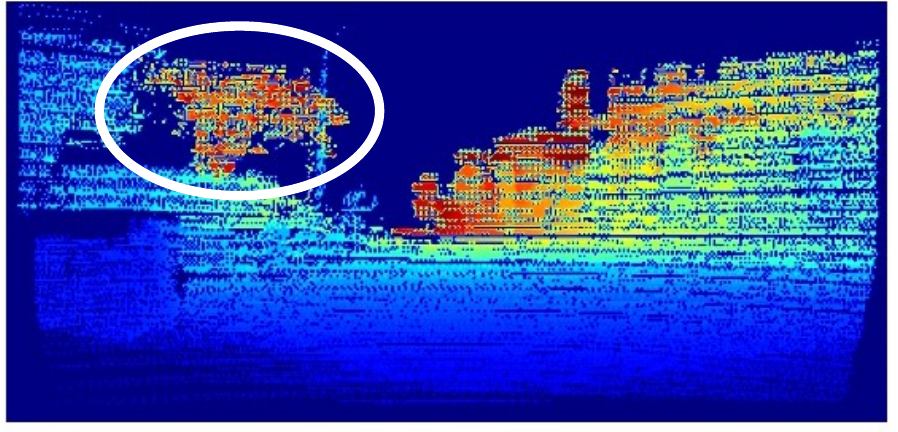}~
    \includegraphics[width=0.32\linewidth]{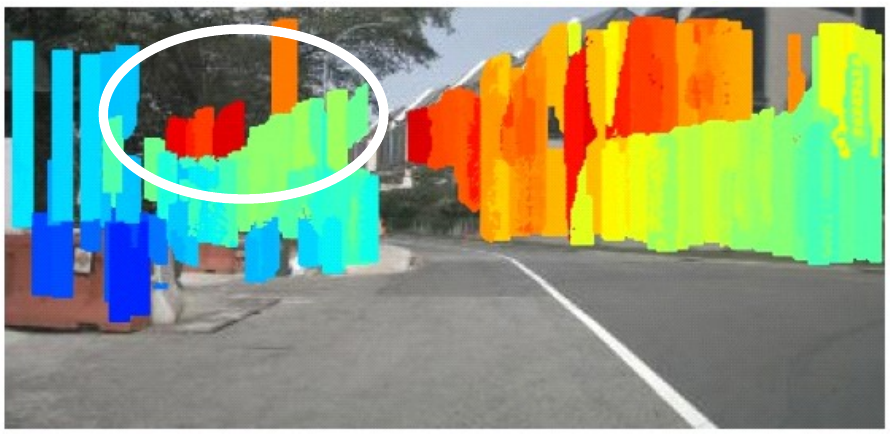}~
    \includegraphics[width=0.32\linewidth]{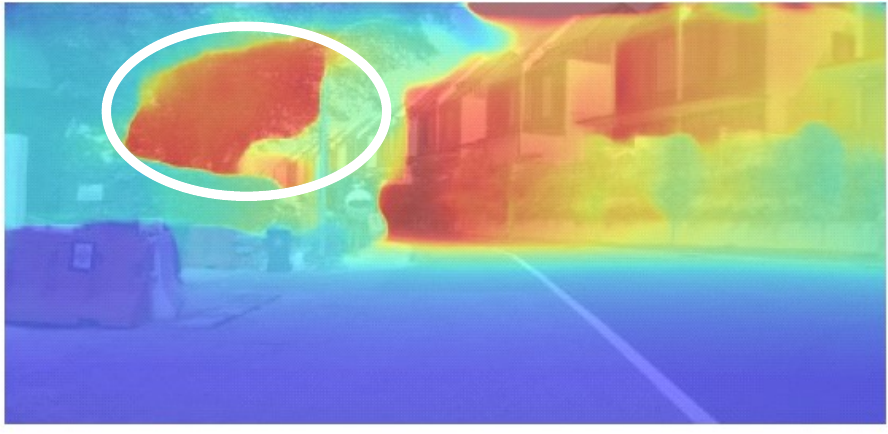}\\
    ~\hspace*{15mm} (b) LM \hspace*{45mm} (d) EM of ours \hspace*{25mm} (f) Depth map completed from (d)
  \end{center}
  \caption{Depth maps. (a) RM. (b) LM. (c) EM of~\cite{DBLP:conf/cvpr/Long00CCN21}. (d) EM of ours. (e) Depth map completed from (c). (f) Depth map completed from (d).}
  \label{fig:depth_estimation}
\end{figure*}

\begin{figure*}[t]
  \begin{center}
     \includegraphics[width=\linewidth]{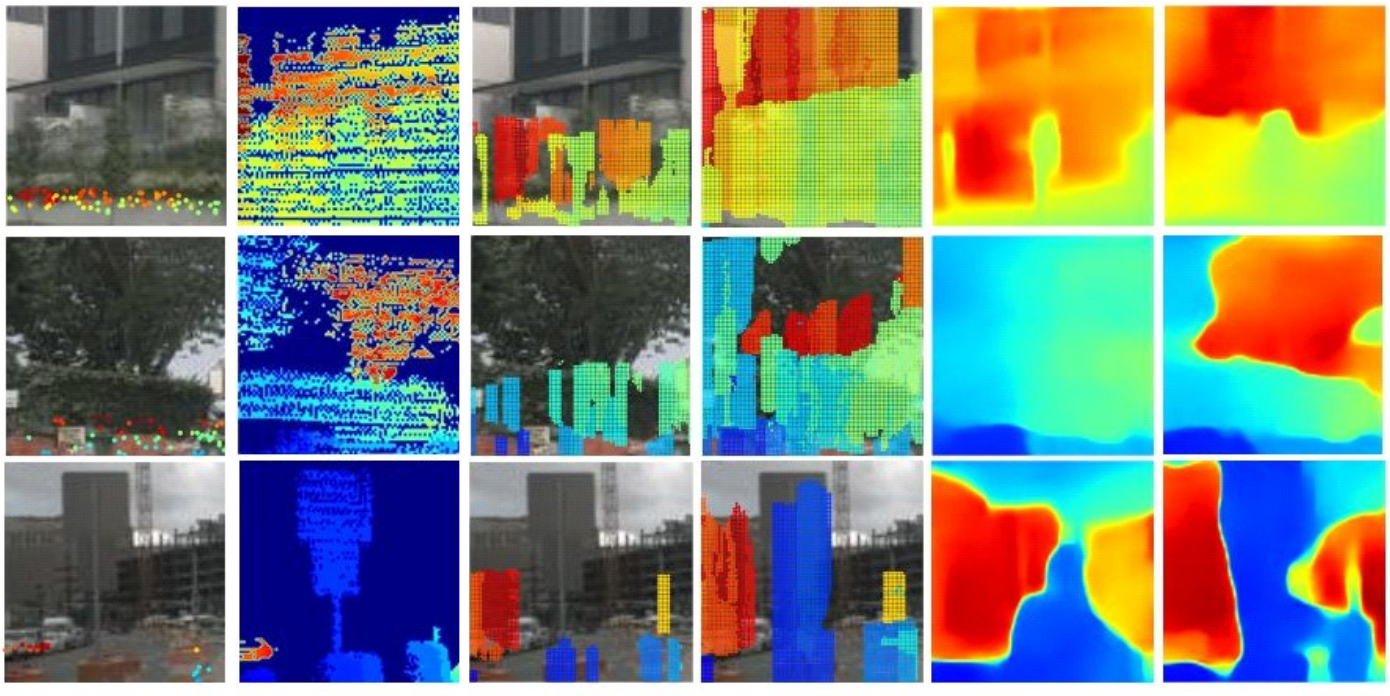}\\
     \vspace*{2mm}
     \begin{minipage}{0.16\textwidth}
     {\center (a) RM}
     \end{minipage}
     \begin{minipage}{0.16\textwidth}
     {\center (b) LM}
     \end{minipage}
     \begin{minipage}{0.16\textwidth}
     {\center (c) EM of the base method~\cite{DBLP:conf/cvpr/Long00CCN21}}
     \end{minipage}
     \begin{minipage}{0.16\textwidth}
     {\center (d) EM of ours}
     \end{minipage}
     \begin{minipage}{0.16\textwidth}
     (e) Depth map completed from (c)
     \end{minipage}
     \begin{minipage}{0.16\textwidth}
     (f) Depth map completed from (d)
     \end{minipage}
  \end{center}
    \caption{Cropped depth maps.}
    \label{fig:qualitative}
    \vspace*{-2mm}
\end{figure*}

\noindent\textbf{Dataset.}
Experiments are conducted with the nuScenes dataset~\cite{DBLP:conf/cvpr/CaesarBLVLXKPBB20} in which RGB images, LiDAR measurements, and radar measurements are included.
Since LiDAR points in this dataset are spatially sparse, they are temporally accumulated over 26 frames with inter-frame spatial alignment using ego-motion in accordance with~\cite{DBLP:conf/cvpr/Long00CCN21}.
The temporally-accumulated depth map is regarded as LM shown in Fig.~\ref{fig:depths} (a).
The image and the depth map are cropped by removing no-depth regions such as the sky, and then downscaled to $400 \times 192$ pixels also in accordance with~\cite{DBLP:conf/cvpr/Long00CCN21}.
The numbers of training, validation, and test images are 12,610, 1,628, and 1,623, respectively.

\noindent\textbf{Parameters.}
Our network is optimized by Adam~\cite{DBLP:journals/corr/KingmaB14} with a learning rate = 5e-5.
The parameters of our method are as follows:
$T_{a} = 1.0$ and $T_{r} = 0.01$.
Since the vertical uncertainty of the millimeter-wave radar used for collecting the nuScenes dataset is at most 20$^\circ$~\cite{radar}, $V$ is set to be 60 under the focal length of the camera.

The depth completion methods~\cite{DBLP:conf/icra/MaK18,DBLP:conf/wacv/LiYLCZZ20} are trained by RMSProp~\cite{DBLP:journals/corr/Ruder16} with momentum = 0.9 and learning rate = 5e-5.

\begin{figure}[t]
  \begin{center}
    \includegraphics[width=\linewidth]{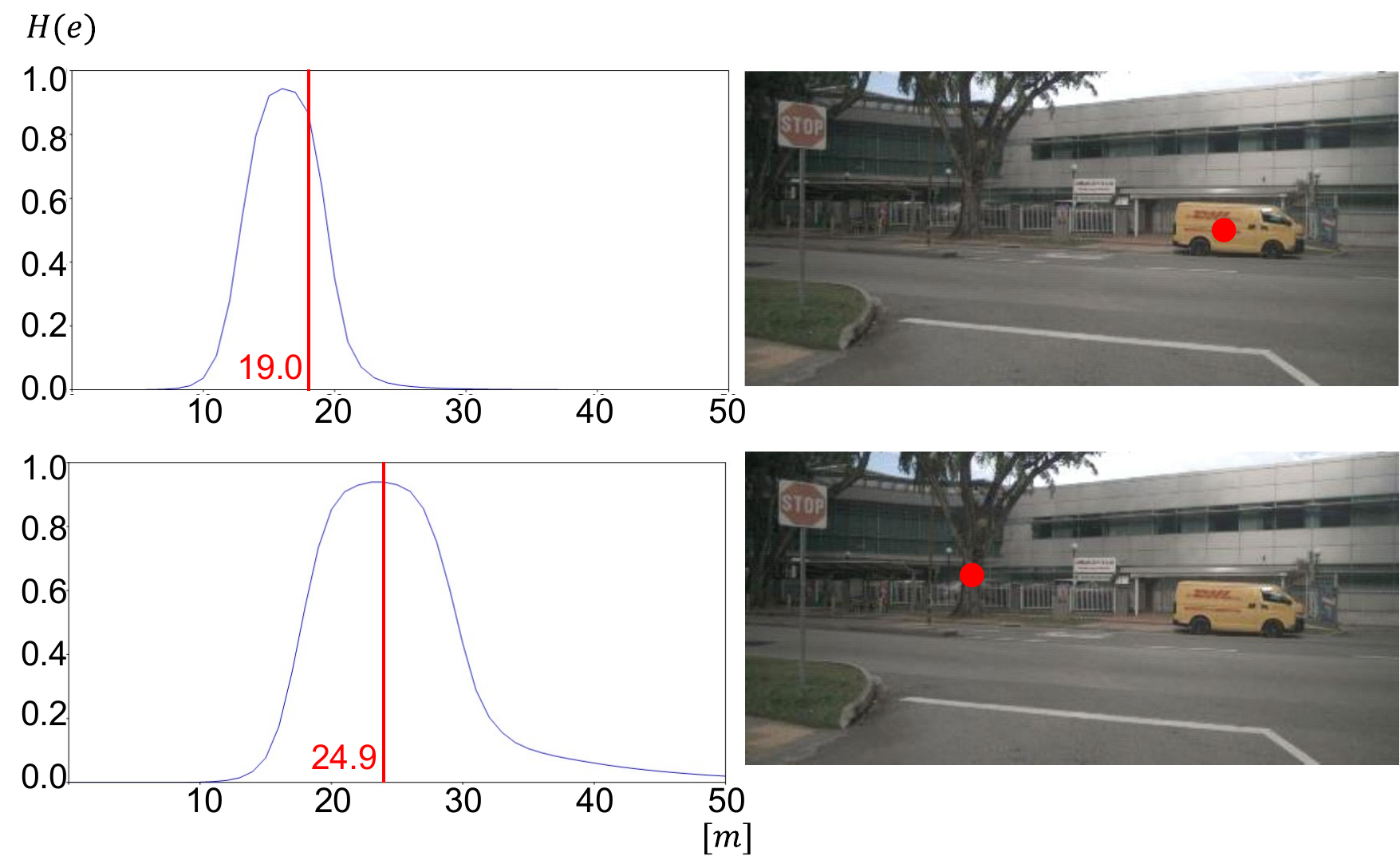}
  \end{center}
  \vspace*{-2mm}
  \caption{(Left) Distribution of $H(e)$. (Right) Pixel of interest indicated by the red point.}
  \label{fig:distribution}
\end{figure}

\subsection{Comparison of Depth Completion Results}

All depth estimation results are evaluated after each sparse depth map estimated by our proposed method (i.e., EM) is completed by depth completion methods (i.e., either of~\cite{DBLP:conf/icra/MaK18} or \cite{DBLP:conf/wacv/LiYLCZZ20}) in accordance with~\cite{DBLP:conf/cvpr/Long00CCN21}.
The completed depth maps are evaluated with the following three metrics: MAE (Mean Absolute Error), RMSE (Root Mean Squared Error), REL (mean absolute RELative error).
For evaluation with these three metrics, the completed depth map is compared with the depth values in LM.
Note that since LM has no-depth pixels, as shown in Fig.~\ref{fig:depths} (a), these metrics are computed only in pixels, each of which has a depth value in LM.

The results are shown in Table~\ref{Tab:result1}.
To validate the performance improvement, the base method~\cite{DBLP:conf/cvpr/Long00CCN21} of our method is also evaluated.
As the baseline and the upper limitation, the depth values of RM and PCRM are completed and evaluated for comparison.
Note that, as mentioned in Sec.~\ref{subsection:preliminary}, (i) RM is the raw radar measurements, and (ii) PCRM is selected from the radar measurements by the supervision of LiDAR points that are not available in inference in real scenarios.

While our method improves the results in all metrics,
the improvement from the base method~\cite{DBLP:conf/cvpr/Long00CCN21} is not significant, and there is a large gap between our method and the upper limitation (i.e., PCRM).
However, the improvements from the baseline (i.e., RM) to the base method~\cite{DBLP:conf/cvpr/Long00CCN21} and our method are 0.10 and 0.13 in MAE, respectively, in the case of depth completion using~\cite{DBLP:conf/icra/MaK18}.
The improvements are ``0.005 and 0.010'' and ``0.14 and 0.26'' in REL and RMSE, respectively, in the case of depth completion using~\cite{DBLP:conf/icra/MaK18}.
This means, while the absolute improvements from the base method are not significant (i.e., 0.03 in MAE, 0.005 in REL, and 0.12 in RMSE), the improvement ratios relative to the base method from the baseline are not minor (i.e., $\frac{0.13}{0.10} = 1.3$ in MAE, $\frac{0.010}{0.005} = 2.0$ in REL, and $\frac{0.26}{0.14} \approx 1.86$ in RMSE).

\subsection{Visualization of Depth Completion Results}

The results of the base method and ours are shown in Fig.~\ref{fig:depth_estimation}.
Given (a) RM and (b) LM, EM is estimated by the base method and ours, as shown in (c) and (d), respectively.
More pixels have depth values in (d) than in (c) because our method expands radar points over $V$ in ERM.
The completed depth maps obtained from (c) and (d) are shown in (e) and (f), respectively.
Different from (c) and (e) of the base method, a distant region enclosed by the white circle is correctly estimated in (d) and (f) of our method; see the same region in (b), which is reliable LiDAR measurements, for reference.

For further evaluation, 
depth maps that are largely different between the base and our methods are shown in Fig.~\ref{fig:qualitative}.
The difference between (c) EM of the base method and (d) EM of ours is observed in higher locations (i.e., the upper sides of the depth maps).
This difference is caused because
possibly correct radar points (i.e., PCRM denoted by $\bm{X}_{p}$) are selected from
the vertically-upward expanded radar points (i.e., ERM) and used for depth map estimation in our method.

While radar point expansion is effective, as mentioned above, it may also spread out erroneous depths.
This negative effect should be suppressed by training negative samples, $\bm{X}_{n}$.
To validate this effect, the distribution of $H(e)$ is shown in Fig.~\ref{fig:distribution}.
As expected, its mean is around the radar depth in both examples.
It is natural that the variance of the distribution is larger in the distant point (i.e., the lower example) due to difficulty in
distant points.
For observing such distant scenes, the effect of Super-Resolution~\cite{DBLP:conf/cvpr/HarisSU18,DBLP:journals/pami/HarisSU21,DBLP:conf/cvpr/ZhangGTSDZYGJYK20,DBLP:conf/eccv/FuoliHGTREKXLXW20,DBLP:conf/iccvw/GuLZXYZYSTDLDLG19} is explored~\cite{DBLP:conf/iconip/HarisSU21,DBLP:conf/ipas2/AkitaHU20,DBLP:journals/access/AkitaU23,DBLP:conf/mva/KondoU21,DBLP:journals/corr/abs-2302-12491}.
Utilizing SR for featurizing distant scenes is an important future direction.


\section{Conclusion}

This paper proposed suppressing the negative effect of the uncertainty of radar directions for depth estimation by fusing image and radar features.
To avoid spreading out the uncertainty over the image, (i) possibly correct radar directions are identified with LiDAR points in training, and (ii) a radar depth is merged with the image features
after feature extraction
in a late fusion manner
for pixelwise depth estimation.


\bibliographystyle{unsrt}
\bibliography{kotani}

\begin{thebibliography}{10}

\bibitem{DBLP:journals/tits/FengHRHGTWD21}
D.Feng et~al.
\newblock Deep multi-modal object detection and semantic segmentation for autonomous driving: Datasets, methods, and challenges.
\newblock {\em {T-ITS}}, 22(3):1341--1360, 2021.

\bibitem{DBLP:conf/wacv/NabatiQ21}
Ramin Nabati and Hairong Qi.
\newblock Centerfusion: Center-based radar and camera fusion for 3d object detection.
\newblock In {\em WACV}, 2021.

\bibitem{DBLP:journals/corr/abs-2205-05335}
J.Hu et~al.
\newblock Deep depth completion: {A} survey.
\newblock {\em CoRR}, abs/2205.05335, 2022.

\bibitem{scanradar}
{Continental}~{Engineering} {Services}.
\newblock {ARS}408-21 {Long} {Range} {Radar} {Sensor} 77{GHz}.
\newblock \url{https://navtechradar.com/clearway-technical-specifications/compact-sensors/}, 2020.
\newblock (Accessed on 3/09/2023).

\bibitem{DBLP:conf/cvpr/RebutOMP22}
J.Rebut et~al.
\newblock Raw high-definition radar for multi-task learning.
\newblock In {\em {CVPR}}, 2022.

\bibitem{DBLP:conf/sdf/NobisGWBL19}
F.Nobis et~al.
\newblock A deep learning-based radar and camera sensor fusion architecture for object detection.
\newblock In {\em SDF}, 2019.

\bibitem{DBLP:conf/cvpr/Long00CCN21}
Y.Long et~al.
\newblock Radar-camera pixel depth association for depth completion.
\newblock In {\em CVPR}, 2021.

\bibitem{DBLP:conf/icra/ChadwickM019}
S.Chadwick et~al.
\newblock Distant vehicle detection using radar and vision.
\newblock In {\em ICRA}, 2019.

\bibitem{DBLP:conf/iros/LinDG20}
J.Lin et~al.
\newblock Depth estimation from monocular images and sparse radar data.
\newblock In {\em IROS}, 2020.

\bibitem{DBLP:conf/wacv/StackerHRS22}
L.St{\"{a}}cker et~al.
\newblock Fusion point pruning for optimized 2d object detection with radar-camera fusion.
\newblock In {\em WACV}, 2022.

\bibitem{DBLP:journals/spl/JiaWLLL21}
Di~Jia, Kai Wang, ShunLi Luo, Tianyu Liu, and Ying Liu.
\newblock {BRAFT:} recurrent all-pairs field transforms for optical flow based on correlation blocks.
\newblock {\em {IEEE} Signal Process. Lett.}, 28:1575--1579, 2021.

\bibitem{DBLP:conf/icra/MaK18}
F.Ma et~al.
\newblock Sparse-to-dense: Depth prediction from sparse depth samples and a single image.
\newblock In {\em ICRA}, 2018.

\bibitem{DBLP:conf/wacv/LiYLCZZ20}
A.Li et~al.
\newblock A multi-scale guided cascade hourglass network for depth completion.
\newblock In {\em {WACV}}, 2020.

\bibitem{DBLP:conf/cvpr/CaesarBLVLXKPBB20}
H.Caesar et~al.
\newblock nuscenes: {A} multimodal dataset for autonomous driving.
\newblock In {\em CVPR}, 2020.

\bibitem{DBLP:journals/corr/KingmaB14}
Diederik~P. Kingma and Jimmy Ba.
\newblock Adam: {A} method for stochastic optimization.
\newblock In Yoshua Bengio and Yann LeCun, editors, {\em {ICLR}}, 2015.

\bibitem{radar}
{Continental}~{Engineering} {Services}.
\newblock {ARS}408-21 {Long} {Range} {Radar} {Sensor} 77{GHz}.
\newblock \url{https://conti-engineering.com/wp-content/uploads/2020/02/ARS-408-21_EN_HS-1.pdf}, 2020.
\newblock (Accessed on 03/09/2023).

\bibitem{DBLP:journals/corr/Ruder16}
Sebastian Ruder.
\newblock An overview of gradient descent optimization algorithms.
\newblock {\em CoRR}, abs/1609.04747, 2016.

\bibitem{DBLP:conf/cvpr/HarisSU18}
Muhammad Haris, Gregory Shakhnarovich, and Norimichi Ukita.
\newblock Deep back-projection networks for super-resolution.
\newblock In {\em CVPR}, 2018.

\bibitem{DBLP:journals/pami/HarisSU21}
Muhammad Haris, Greg Shakhnarovich, and Norimichi Ukita.
\newblock Deep back-projectinetworks for single image super-resolution.
\newblock {\em {IEEE} Trans. Pattern Anal. Mach. Intell.}, 43(12):4323--4337, 2021.

\bibitem{DBLP:conf/cvpr/ZhangGTSDZYGJYK20}
Kai Zhang et~al.
\newblock {NTIRE} 2020 challenge on perceptual extreme super-resolution: Methods and results.
\newblock In {\em CVPRW}, 2020.

\bibitem{DBLP:conf/eccv/FuoliHGTREKXLXW20}
Dario Fuoli et~al.
\newblock {AIM} 2020 challenge on video extreme super-resolution: Methods and results.
\newblock In {\em ECCVW}, 2020.

\bibitem{DBLP:conf/iccvw/GuLZXYZYSTDLDLG19}
Shuhang Gu et~al.
\newblock {AIM} 2019 challenge on image extreme super-resolution: Methods and results.
\newblock In {\em CVPRW}, 2019.

\bibitem{DBLP:conf/iconip/HarisSU21}
Muhammad Haris, Greg Shakhnarovich, and Norimichi Ukita.
\newblock Task-driven super resolution: Object detection in low-resolution images.
\newblock In {\em ICONIP}, 2021.

\bibitem{DBLP:conf/ipas2/AkitaHU20}
Kazutoshi Akita, Muhammad Haris, and Norimichi Ukita.
\newblock Region-dependent scale proposals for super-resolution in object detection.
\newblock In {\em IPAS}, 2020.

\bibitem{DBLP:journals/access/AkitaU23}
Kazutoshi Akita and Norimichi Ukita.
\newblock Context-aware region-dependent scale proposals for scale-optimized object detection using super-resolution.
\newblock {\em {IEEE} Access}, 11:122141--122153, 2023.

\bibitem{DBLP:conf/mva/KondoU21}
Yuki Kondo and Norimichi Ukita.
\newblock Crack segmentation for low-resolution images using joint learning with super- resolution.
\newblock In {\em MVA}, 2021.

\bibitem{DBLP:journals/corr/abs-2302-12491}
Yuki Kondo and Norimichi Ukita.
\newblock Joint learning of blind super-resolution and crack segmentation for realistic degraded images.
\newblock {\em arXiv}, 2302.12491, 2023.

\end{thebibliography}

\end{document}